# On-Orbit Smart Camera System to Observe Illuminated and Unilluminated Space Objects

**Steven Morad**
*Space and Terrestrial Robotic Exploration Laboratory, University of Arizona*
**Ravi Teja Nallapu**
*Space and Terrestrial Robotic Exploration Laboratory, University of Arizona*
**Himangshu Kalita**
*Space and Terrestrial Robotic Exploration Laboratory, University of Arizona*
**Byong Kwon**
*Space and Terrestrial Robotic Exploration Laboratory, Arizona State University*
**Vishnu Reddy**
*Lunar and Planetary Laboratory, University of Arizona*
**Roberto Furfaro**
*Systems and Industrial Engineering, University of Arizona*
**Erik Asphaug**
*Lunar and Planetary Laboratory, University of Arizona*
**Jekan Thangavelautham**
*Space and Terrestrial Robotic Exploration Laboratory, University of Arizona*

## ABSTRACT

The wide availability of Commercial Off-The-Shelf (COTS) electronics that can withstand Low Earth Orbit conditions has opened avenue for wide deployment of CubeSats and small-satellites. CubeSats thanks to their low developmental and launch costs offer new opportunities for rapidly demonstrating on-orbit surveillance capabilities. In our earlier work, we proposed development of SWIMSat (Space based Wide-angle Imaging of Meteors) a 3U CubeSat demonstrator that is designed to observe illuminated objects entering the Earth's atmosphere. The spacecraft would operate autonomously using a smart camera with vision algorithms to detect, track and report of objects. Several CubeSats can track an object in a coordinated fashion to pinpoint an object's trajectory. An extension of this smart camera capability is to track unilluminated objects utilizing capabilities we have been developing to track and navigate to Near Earth Objects (NEOs). This extension enables detecting and tracking objects that can't readily be detected by humans.

The system maintains a dense star map of the night sky and performs round the clock observations. Standard optical flow algorithms are used to obtain trajectories of all moving objects in the camera field of view. Through a process of elimination, certain stars maybe occluded by a transiting unilluminated object which is then used to first detect and obtain a trajectory of the object. Using multiple cameras observing the event from different points of view, it may be possible then to triangulate the position of the object in space and obtain its orbital trajectory. In this work, the performance of our space object detection algorithm coupled with a spacecraft guidance, navigation, and control system is demonstrated.

In our tests, we were able to successfully detect a transit 88% of the time with $\sigma = 0.5$ DN sensor readout noise. Our method scales linearly in time and with the number of pixels, with the most computationally intensive phases being parallelizable and simple enough to be offloaded to SWIMSat's onboard FPGA. A thorough description of the detection algorithm, along with the tracking controller is presented in this work. Our work suggests both a critical need and the promise of such a tracking algorithm for implementation of an autonomous, low-cost constellation for performing Space Situational Awareness (SSA).

## 1. INTRODUCTION

Planet Earth experiences several kinds of space hazards on a continuous basis. One is the impact of meteoritic debris in the form of dust and small pebbles (estimated to be ~100,000 tons a year) [1]. Another is the impact of meter to kilometer sized objects that can have catastrophic impact. The small dust and pebbles travel at speeds of up to 70 km/s coming from various sources including passing comets, asteroids, the Moon, nearby planets and even remains of interstellar dust. Much of these pebbles and dust particles ablate, vaporizing or shattering in the earth's atmosphere. However, in space, these particles pose a serious hazard to satellites and human spacecraft. The flux of this debris has varied greatly in time, and may be experiencing a considerable upturn in the present day [2]. Attempts to begin detection and tracking of these objects as they enter the atmosphere can provide meaningful data of their source, composition and threat magnitude.

Objects ~1-50 m diameter may be impacting an order of magnitude more frequently than usual, perhaps due to recent breakups or other dynamical events [2]. Most recent of the larger space rocks is the ~17 m diameter LL5 chondrite that exploded over Chelyabinsk, Russia in Feb. 2013, from which a 0.5-ton fragment was recovered [3-4].

Over the 20-year interval 1994-2013, US government sensors recorded at least 556 bolide events of various energies, as shown in Fig. 1 (from neo.jpl.nasa.gov). The times (day/night) and energies of events are pictured by orange or blue circles respectively. The largest dots, around 100,000 GJ, correspond to ~20 kT explosions. Most of the energy of such events goes into the fireball, the ablating trail, the sonic boom, and the vaporization of the material. The most immediately visible part of a fireball is the impact flash itself, which can last for seconds to tens of seconds depending on the size and the impact angle preceded by the ablating trail. Assuming, conservatively, that only 3% of the impact energy goes into the visible flash, then the visible peak brightness of a 1 m class event, with a kinetic energy of 0.1 kT, is about equal to a thousand lightning strikes going off at once [2].

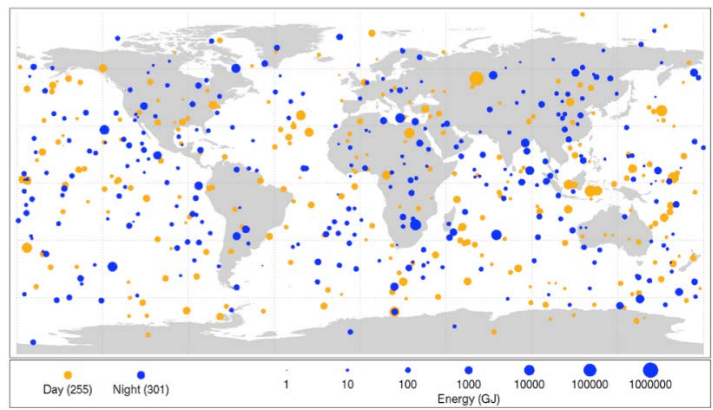

Fig. 1. Meteor impacts events totaling more than 500 from 1994 to 2013 recorded using US government sensors.

The rapid miniaturization of electronics, sensors, actuators and exponential growth in computational capability of microprocessors all have enabled small-spacecraft such as CubeSats. For the cost of several-hundreds of thousands of dollars, it is now possible to launch several space observatories that avoid the limits of ground observatories from occlusion, weather and atmospheric distortion. Utilizing several space observatories, it is possible to perform multipoint observation that enable measuring the kinetics of moving objects such as meteors. Using these advances, we hope to bridge the theoretical with the observational and computational sciences to get a clearer picture and dangers of the meteor and Near Earth Object phenomena. In this paper, we propose an autonomous and efficient method of detecting and tracking meteors and NEOs that do not emit enough light to be visible to CubeSat-grade visible-light sensors. Importantly, using these techniques, the spacecraft can detect and track meteors well before they enter the ablation phase in the atmosphere. In the following sections, we present background and related work, followed by a system overview of the SWIMSat mission, followed by presentation of the unilluminated object detection algorithm, results and discussions.

## 2. BACKGROUND AND RELATED WORK

An important source of meteors are particles released from comets during their perihelion passage, or remains of asteroids upon collision. Some of these meteors are considerable in size and occur regularly in the form of annual meteor showers. Meteors also trace their origin to the Moon and nearby planets, escaping these bodies from an impact event. Other meteor traces their origin to Near Earth Objects (NEOs), the main asteroid belt, the Kuiper belt and to inter-stellar dust. Meteor more than 1 mm are known to have enough surface to mass ratio to ablate through the earth's atmosphere. Large meteors can survive (Fig. 2) the journey to Earth's surface and are referred to as meteorites.

The meteoroid as it enters the Earth's atmosphere losses mass due to sputtering due to high-energy collision with rarefied atmospheric gases [5-7]. As the meteor then enters exponentially increasing atmospheric density, it undergoes rapid heating and may undergo ablation [8]. Ablation occurs only if the particle reaches sufficiently high temperatures. Combination of the high temperatures and atmospheric forces may cause fragmentation. Most particles in the size range of 0.1 – 10 cm ablate at 70 and 100 km altitude. It is therefore of utmost importance to be able to fully track a meteor entering Earth before it starts to ablate. This enables more accurate estimates of energy released.

Conventional methods for meteor and Near Earth Object (NEO) detection involve identify streaks as the meteor is ablating. Shin et al. uses a RAndom SAmple Consensus (RANSAC) method to obtain a linear trajectory for NEOs from visible streaks [9]. Another more sophisticated approach is called transit photometry and has been widely used to discover exoplanets. This same technique is being demonstrated here to detect Near Earth Objects and unilluminated meteors.

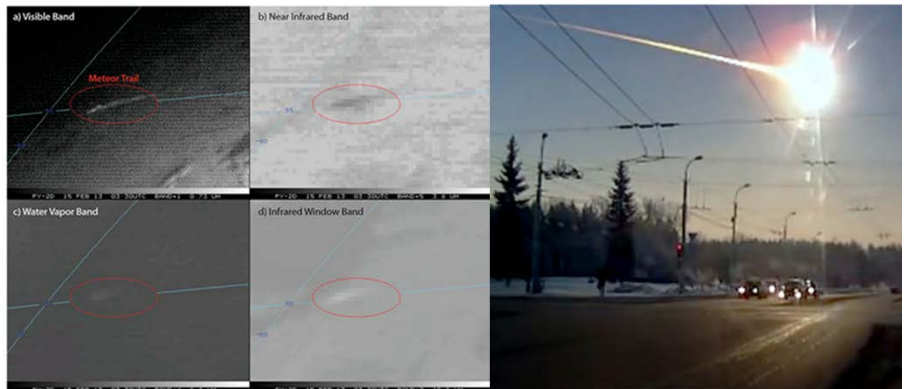

Fig. 2. (Left) Chelyabinsk Meteor trail first detected by Chinese Geostationary satellite, Feng-Yun 2D (positioned at 86.5°E) 12 min after trail formation. (Right) Observed meteor entry from ground.

McInroy et al. discuss visual silhouetting as a method for tracking via boundary generation [10]. The boundary expands by adding thick line segments to the boundary, while we fit a single line.

In our approach, we make a simple assumption, expecting the entire transit to appear as a single line, but it is our understanding that by limiting the trajectory to a single line increases robustness to sensor noise. In contrast, McInroy et al. [10] method requires the shape of the object being detected and tracked known beforehand, as the initial search phase fits a polygon to the image. This is not directly applicable to meteors as they appear with no forewarning.

Ground observation of meteors are affected by limited field of view of the observing instruments. As a result, the event may not be fully observed from the ground. Complete observations of the event can provide a holistic understanding of the origin, the evolution and end of a meteor as it enters the earth's atmosphere. Being able to observe the fully evolution of the meteor entry can give tell-tale clues of compositional differences between cometary and asteroid meteors. An additional challenge to ground based observations is the atmospheric disturbances, and occlusions. The procedures and challenges for ground based observations are described in [11]. These challenges can be overcome by high-altitude or space based observations of meteors. Notable high-altitude observations of the Leonid shower were conducted in 1998 [12] and 1999 [13], where the observations were made from an aircraft equipped with cameras, LIDARS, and spectrographs. The success of this mission provided scope for successor missions [14-15]. However, aircraft missions are specifically timed to a meteor event, and hence cannot be the solution for long-term, large-scale meteor monitoring.

There lacks a dedicated satellite network to observe and characterize meteor impacts in the upper atmosphere. This would require selection of camera with right wavelength to pick up hot meteor trails and other distinct characteristics of meteor impacts, and be optimized to detect and track fireball clouds that persist for hours after the explosion. Current data is gathered from other satellites that happen to catch a glimpse of a meteor event. Long, dedicated observation time will help to quantify the true effect of these meteor impacts onto Earth, their frequency, size and some basic characterization regarding the meteor trail that is created [4]. We are developing SWIMSat as prototype CubeSat nodes to enable detection and tracking of meteors from Low Earth Orbit (LEO). In the following section, we present an overview of the SWIMSat mission concept followed by the vision algorithm.

## 3.   SWIMSAT MISSION CONCEPT

The proposed spacecraft design has matured thanks to a Phase A/B design study contract with the US Airforce through the University Nanosatellite Program (UNP). The contract enabled the design team to bring the design from an early concept to a refined design approaching Critical Design Review [16-18]. Significant development work occurred that identified the major contributions in the project including selection of the right cameras for wide angle and narrow angle observation, the right computer interface to perform timely autonomous detection of the meteor events, in-addition to development of meteor detection algorithms. All subsystems achieved brass-board maturity (i.e. early working prototypes that can be tested in the laboratory under controlled conditions).

## 3.1 SPACECRAFT

The proposed SWIMSat nodes consists of two CubeSats, each a 3U (10 cm × 10 cm × 34 cm) spacecraft (Fig. 3) with a mass of 4 kg. The pair of spacecraft will be located in Low Earth Orbit and will be able to observe meteors within a 200 km diameter area. Each spacecraft will be at the same altitude but be at approximately 25 km separation distance (determined during deployment). Each CubeSat will be equipped with two science imagers, namely a Wide-Angle Camera (WAC) and a Narrow Angle Camera (NAC). Each camera will use Sony CMOS detectors that can be set to 1.3 to 36 Megapixels resolution.

Each CubeSat will use the WAC to autonomously scan for meteor events. Once a meteor is detected, the spacecraft will instantly rotate 90º to point the NAC and continue to track the meteor event. The NAC will enable zoomed observation of the meteor. Upon autonomous detection and tracking by one of the CubeSats, the second CubeSat will be called by the first CubeSat to track the same meteor. Once the two CubeSats continue to monitor the incoming object, they will be able to determine object position, velocity, deceleration, and angle of entry into the atmosphere. Detailed event reports will be transmitted down to ground in real-time.

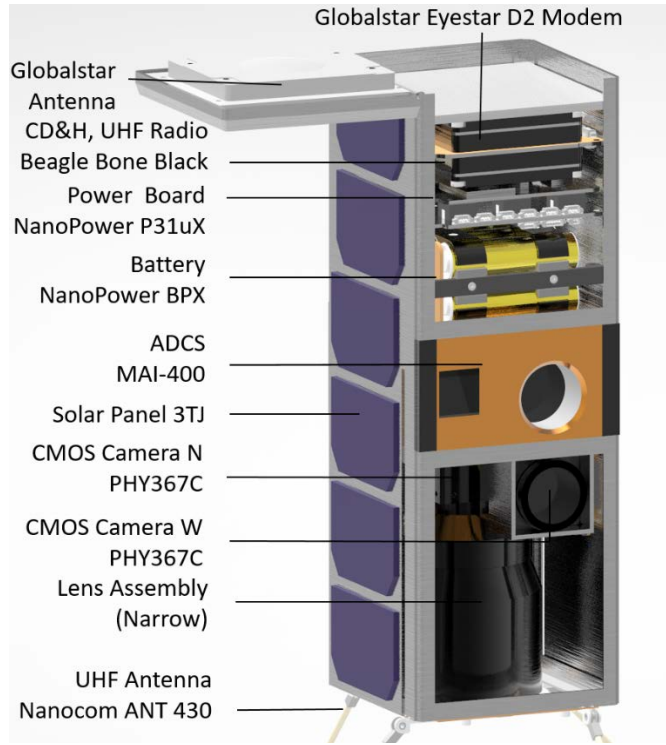

Fig. 3. SWIMSat CubeSat Node Layout.

Each spacecraft will be powered using body-mounted Spectrolab solar cells containing triple junction cells providing an average of 7 W of power during daylight. The system will charge a 78 Whr GOMSpace Lithium Ion battery. Depth of discharge will not exceed 50 % to maximize battery capacity and life. Power subsystem design suggests that it is sufficient for mission needs (see Table 1). The on board Attitude Determination Control System (ADCS) consists of the Maryland Aerospace MAI-400 that combines a suite of sensors such as two horizon sensors, MEMS IMU, 3 reaction wheels and 3 magneto-torquers. The system permits pointing at 1-2º with 3-axis stabilization.

This unique 3U CubeSat design provides robust system margins (see Table 1). Note the budget includes 10% component margin. The spacecraft uses a ISIS Command and Data Handling (C&DH) computer board that contains an extremely power-efficient ARM 9 processor, with spacecraft watchdog functions and overall control of navigation, communication and control. The ISIS CD&H has 6-levels of watchdog that enable robust handling of Single Event Upsets (SEUs). The spacecraft will use the Beaglebone Black combined with the Beagle Bone Cape containing a Xilinx Spartan 6 LX9 FPGA board for rapid image processing. The Beagle Bone Black + Cape has been flown on the RadSat CubeSat mission and is TRL 9.

Transmitting down all of the captured video will heavily tax the spacecraft due to the high power required for data transmission. This also presents challenges for thermal control. In addition, it will require ground operators to be constantly on-watch which is logistically complex. Therefore, these challenges require the spacecraft software perform data reduction and enable autonomous detection, tracking and reporting of critical events. Beagle Bone Black with Black Cape FPGA board is well designed for this task, especially because it can be used to easily parallelize computational processes.

The spacecraft will utilize NSL EyeStar-D2 Duplex Globalstar and UHF/VHF radio system for communication. The Globalstar radio permits near-real-time communication with the spacecraft no matter where it is orbiting the planet. The radio has a data rate of 700 Bytes/second which adds up to 60 MB over 24 hours. This is comparable to a S-band transmitter operating at 3 Mbits with transmit time of 3 minutes per day. However, this avoids the cost of setup and maintaining an S-band ground station.

Electronics and all other thermal sensitive components will be located on top of the craft. High heat producing components are isolated and are located well away from other temperature sensitive electronics. The heat from internal components (using a metabolic heating strategy), in combination with resistive heaters will maintain a temperature of 0 to 45 ºC. Critical thermal vacuum cycling tests will be performed to verify the as-built thermal design. The spacecraft will contain a deployable radiator to radiate excess heat from the Globalstar radio into space.

The proposed spacecraft design borrows many elements from AOSAT I (CubeSat centrifuge mission to operate in Low Earth Orbit) [19-21] that will be launched in 2019, including use of common bus components, computer, power system and attitude control software.

**Table 1.** SWIMSat Node Systems Budgets

| System | Mass (kg) | Volume (cm$^3$) | Avg. Power (W) |
|---|---|---|---|
| Communications | 0.3 | 500 | 0.5 |
| Onboard CPUs | 0.2 | 50 | 2 |
| Instruments | 0.9 | 450 | 2 |
| Power Conv. | 0.45 | 600 | 0.5 |
| Attitude Det. & Control | 0.76 | 500 | 2 |
| Structure | 0.45 | 250 | - |
| Thermal | 0.1 | 50 | 0.1 |
| Total | 3.0 | 2400 | 7.1 |
| Margin | 22 % | 20 % | 14 % |

### 3.2 CONCEPT OF OPERATIONS

A concept of operations for the proposed spacecraft is shown in Fig. 4. The spacecraft will be launched into a Low Earth Orbit (LEO) of 420 to 650 km altitude on a 12-month primary mission. As a baseline, we presume the CubeSats will be deployed from the International Space Station. The first month will be spent calibrating the instruments and testing all subsystems to ensure the system is fully operational. This will include collective calibration between the two CubeSats to pinpoint position of objects being tracked by both spacecraft. After two months, the pair of CubeSats will be ready to perform monitoring and tracking of meteor and NEO events. This will give us an opportunity to fully test the detection algorithms.

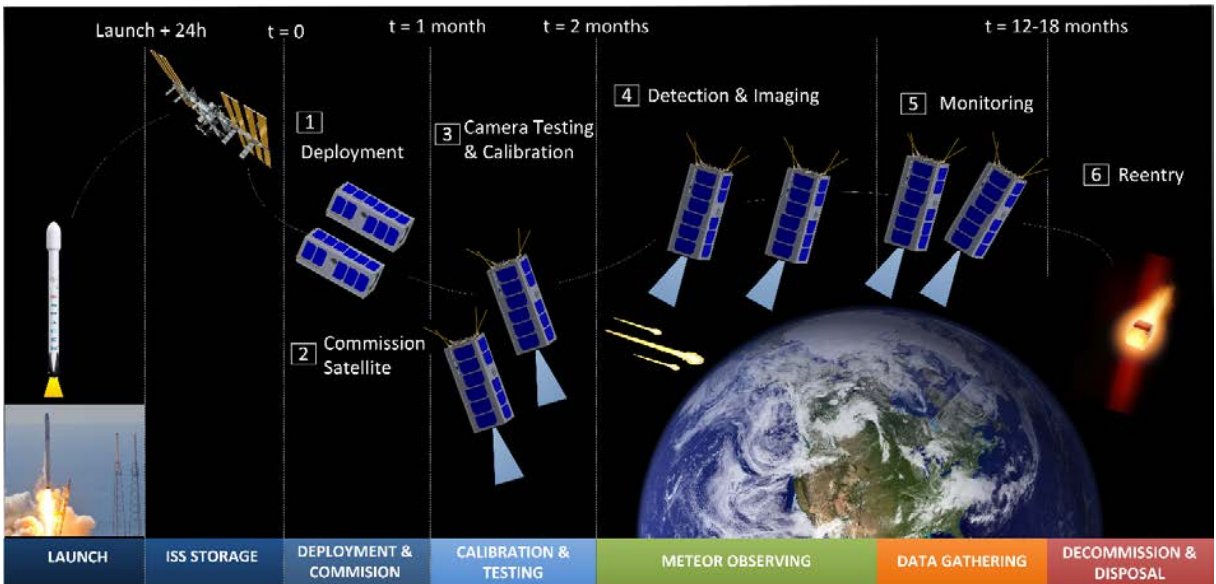

Fig. 4. SWIMSat Concept of Operations

The spacecraft upon autonomous detection of meteors and NEOs will send thumbnail pictures and critical data of the event by using the Globalstar radio. This will be followed by tracking video and higher quality still-shots of the event from start to finish or when the CubeSats loses track of the event. Critical data about each event will be published including magnitude, position, entry angle into the atmosphere, velocity, acceleration and estimate of impact altitude.

Along the way, with operational experiences, improvements will be made to reduce false positives with the detection algorithm. The collected data will be compiled into datasets available to the science community for evaluation with other software. After 12-18 months the spacecraft will have entered a fast degrading orbit and will undergo mission disposal. In the following section we present the autonomous object detection algorithm.

## 4. OBJECT DETECTION ALGORITM

The goal of the object detection software is to (1) detect (2) image (3) track incoming meteors and transiting NEOs. The detection and tracking software will be resident on the spacecraft, enabling each CubeSat to perform this activity autonomously. The software architecture is shown in Fig. 5. In this architecture, raw images undergo filtering and masking followed by application of a feature detection algorithm to detect meteors. Once detected, the software will then call upon the neighboring CubeSat and itself to start tracking the event. Once the event is tracked, estimates of the position, velocity, acceleration and size of objects will be calculated using techniques outlined in Section 5.1. This will culminate with compilation of videos and images that will be communicated to ground.

Several different techniques are used for detection and tracking. This includes blob detection [16] (Fig. 6), optical flow [16, 18], and Hue-Saturation-Value (HSV) filtering [18] with plans to use neural networks that will train on sets of meteor images developed using a physics based simulator. To date the blob detection method has provided 70-90% accuracy in correctly detecting meteor events using both real and artificial data sources. The blob detection method uses filters to first find a bright tail, followed by using blobs to detect the head. Once a threshold number of blobs are found that match this shape, then a 'rubber band' is drawn around the identified meteor event.

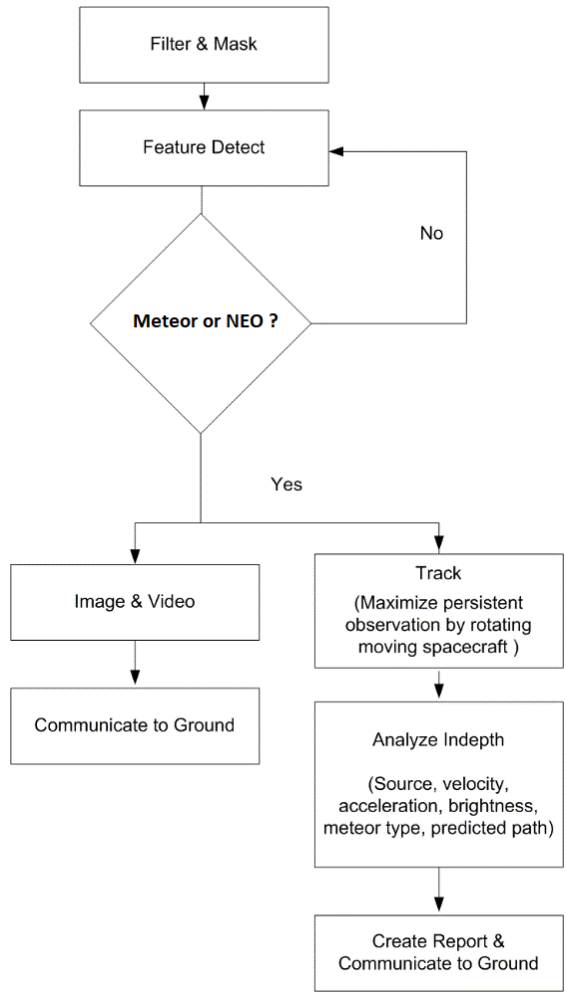

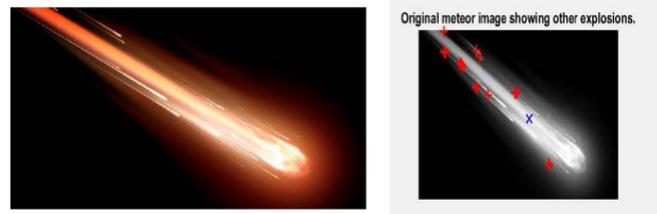

Fig. 6. Blob detection identifies collection of bright pixels in prescribed shape, of length to width ratio.

Fig. 5. Autonomous Object Detection Software Architecture

Our proposed NEO and meteor detection method for unilluminated objects is inspired by transit photometry. Transit photometry is where bodies orbiting a star cause a dip in the light flux as they pass between the star and the observer. This method is generally used for detecting and characterizing exoplanets [22]. We apply this mode of thinking to the entire night sky. We track each visible star in the star field, and look for changes in the flux as a NEO passes between the star and our observer. With the smaller instruments one would find on a CubeSat, the star field tends to be sparse. However, given enough time, we are able to collect enough data to infer the existence of a NEO and compute its trajectory.

This algorithm operates on two temporally ordered images at a time, image *n* and image *n+k*, for k>0 which we refer to as the reference image and operating image respectively. First, the stars need to be categorized so decreases in flux can be tracked. There are many star-trackers that categorize stars in the star field, many with preprogrammed maps of the sky. For simplicity, we devised our own system based on Otsu's Binarization. We use Otsu's Binarization to categorize each pixel of an image as bright or dim. Otsu's Binarization is a nonparametric algorithm used in computer vision to sort a bimodal distribution of values into two distinct types. We run Otsu's Binarization before any operations on both our reference and operating image, to greatly reduce the search space of our method. The position of each bright pixel corresponds with a star and has its properties recorded separately for each image.

After correcting for parallax shift between the two images, the operating star map can be subtracted from the reference star map. This operation can be done in parallel for each star, leveraging the use of the FPGA present on SWIMSat. The result is a set of anomalies, pixels that were bright in the reference image but dark in the operating image. This step inherently generates many false positives due to sensor readout noise and drift.

Anomalies are stored for a period of *j* successive images, where *j>k*. The reference and operating images alone do not contain enough data to detect a transiting NEO. As we approach the *j*th image uncertainty begins to decrease. We know that sensor readout noise is Gaussian per-pixel, and that generally this noise is independent across all pixels (barring sensor defects). This implies the noise we see in each image is uniformly distributed across all the pixels. The trajectory of the NEO we expect to see are linear, as shown below. This allows us to extract useful data, even when the S/N ratio is below 1.

We show that the objects we track appear linear to our observation satellite by taking the partial derivative of Kepler's equation of motion:

$$r = \frac{a(1-e^2)}{1+e\cos f} \qquad (1)$$

where:

$$\frac{dr}{df} = \frac{a(1-e^2)e\sin f}{(1+e\cos f)^2} \qquad (2)$$

Eccentricity *e* and semi-major axis *a* of the object are constant during an observation, only true anomaly *f* is changing. Although *f* varies, changes in *f* are very small for the timescale of our observation. This allows us use the small angle approximation for sin *f* and cos *f*, making *dr/df* linear. This implies that the path of a NEO will appear linear to our observer. Now we can use a simple linear model to detect and track the object.

We capitalize on the apparent linearity of the trajectory by using RANdom SAmple Consensus (RANSAC) to generate a linear model from the aggregated anomalies. RANSAC operates very well on datasets with uniformly distributed outliers [23]. Like the star map subtraction, RANSAC lends itself very well to parallelism, and has already been implemented for FPGAs [24]. The model loss *L* of the model is used to classify whether or not a detection has occurred. When *L* is below a threshold, the resulting linear model is used as the estimated trajectory of the object.

The presented algorithm is composed of three computationally expensive parts, the binarization phase, the star search phase, and the RANSAC phase. Otsu's Binarization runs in $O(255 + n)$ where *n* is the number of pixels [25]. The star search phase does a Boolean check of every pixel which evaluates to $O(n)$. The RANSAC phase consists of running *R* linear regressions with *S* stars for a runtime complexity of $O(4RS)$. Computing the loss of inliers takes $O(2N)$, where *N* is the subset of stars that contribute to the line fit. Combining these results in $O(k\,(4RS + 2N)\,)$ [14], where *k* is the user-set upper limit on number of iterations. The default value of *k=100* worked for us. The number of stars *S* is a small fraction of the number of pixels, so the time complexity of the RANSAC phase is dominated by the other phases. The overall runtime complexity of our method scales linearly with the number of pixels.

## 5. RESULTS

Tests were run on a synthetic 2000×1000-pixel star field displaying stars with an apparent magnitude of 6.5 or less, (Fig. 7) from [26]. Uniformly distributed readout noise was applied to the star field with a normal distribution of σ = 0.5 DN per pixel. Otsu's Binarization began to break down around σ = 0.6 and limited us from adding more noise. In the future, this could be replaced by a Binarization method developed specifically for star fields. A 8-bit pixel intensity range was used, resulting in roughly σ = 0.2 of the total pixel range. The values *k=1* and *j=30* were used as our algorithm parameters.

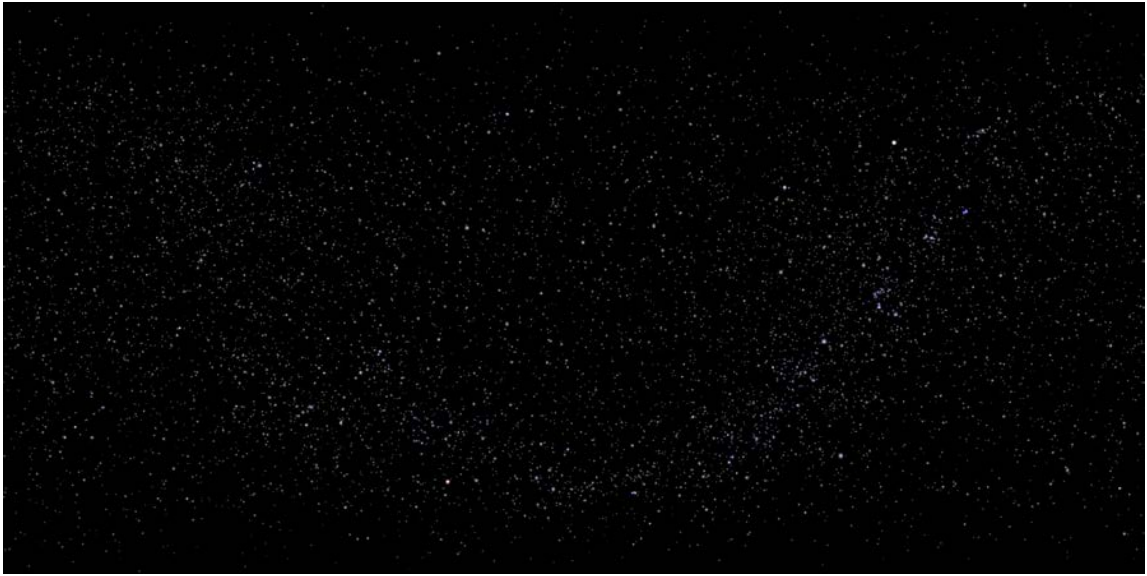

Fig. 7. A synthetic starfield of < 6:5 magnitude stars.

Linear NEO trajectories were randomly generated, with start and end points along $x=0$ and $x=2000$ respectively. The $y$ values were uniformly selected from the domain [0, 1000]. The position of the NEO transit was simulated over 30 frames (Fig. 8).

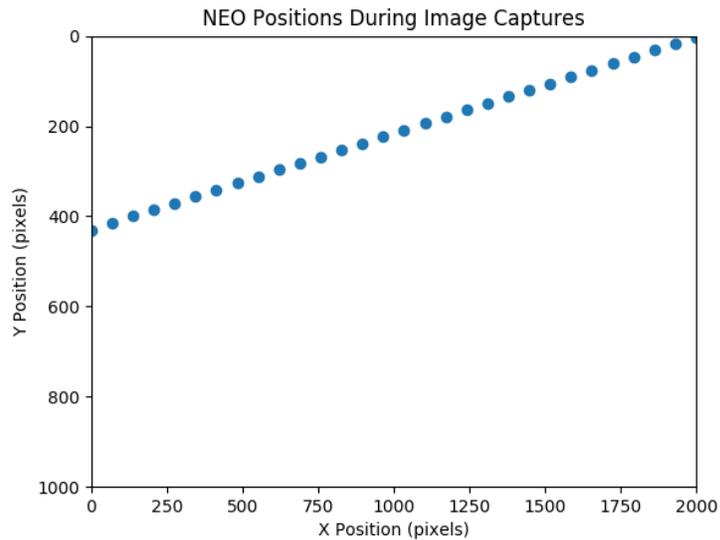

Fig. 8. An NEO's position compiled from a series of images superimposed. Note that the dots are not the actual size of the occultation, they are greatly enlarged to aid the reader.

The occultation of the NEO was simulated with a 3-pixel radius black circle (Fig. 9). Over non-star pixels, the algorithm failed to detect anomalies from the object, as expected. In fact, the NEO only passed over a bright pixel in roughly 30% of the total frames (Fig. 10 left). 25 trials were run with these parameters, of which 22 were flagged as detections by the algorithm. The standard deviation between the actual trajectory and the estimated trajectory was 6.15 pixels (Fig. 10 right). On an 8-core 1st generation AMD Ryzen processor, the mean processing time per frame was 0.97 seconds. By rewriting the program in C and offloading the model generation and star map operations to a SWIMSats FPGA processor, we believe the current iteration is efficient enough to run on the SWIMSat spacecraft.

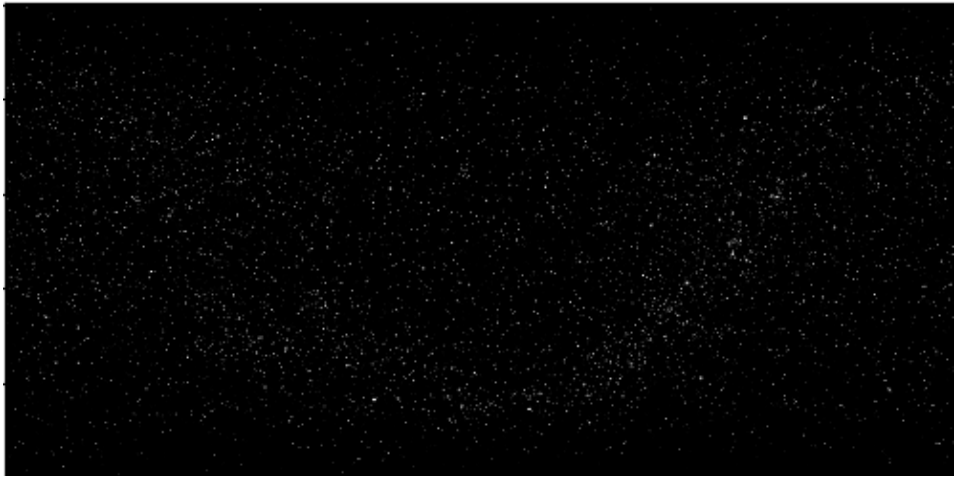

Fig. 9. The star field with a NEO of radius of 3 pixels transiting over 30 frames superimposed on top. Human detection and tracking of these unilluminated objects is impossible.

The noise models here could be improved in the future. Pointing noise was not taken into account due to the difficulty in generating an accurate model for CubeSats. Ultimately, another step will exist to filter pointing noise and to reorient the frame to account for pointing drift. This paper focuses on what a single satellite can do. However, for three dimensional localizations another observation of the NEO transit is required. This observation could be a ground-based asset, but it would be interesting to see how multiple SWIMSats could work together to localize and track a NEO. Multiple SWIMSat observations could reduce the trajectory error and improve detection rates.

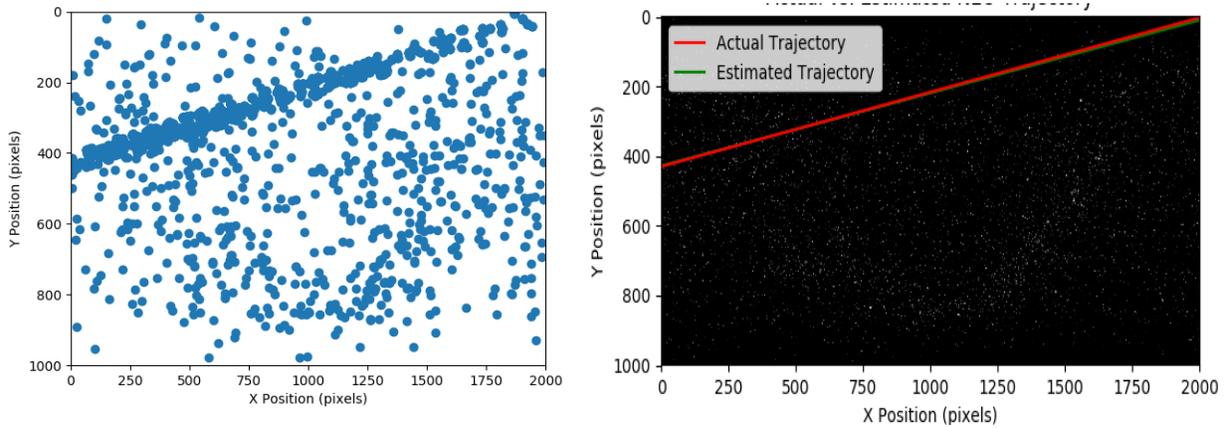

Fig. 10. (Left) The anomalies are detected over many frames and plotted. The simulated Gaussian sensor readout noise makes fitting a line more difficult. At σ = 0.5 DN and below, the binarization process filters out much of the noise. However, with larger sigmas, Otsu's binarization begins to break down. (Right) One typical example of the actual trajectory of the NEO and the estimate produced by our algorithm

### 5.1 MULTIPOINT OBSERVATION

Here we describe multipoint observation using 2 spacecraft. Using 2 GPS equipped spacecraft it is possible to concurrently track another object and determine its position. Using the position calculation, we may then extend this to calculating velocity and acceleration through change in time. This capability can be easily extended to *n* spacecrafts to cover a greater range or obtain increased accuracy. Now consider the image is seen by 2 spacecrafts A and B as shown in Fig. 11.

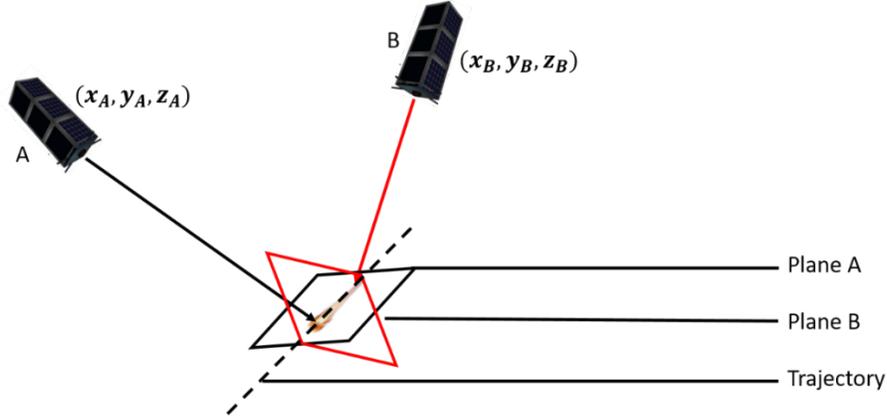

Fig. 11. Viewing geometry of 2 spacecrafts, where the intersection of the viewing planes yields the trajectory.

Let the vectors $[a_A, b_A, c_A]$ and $[a_B, b_B, c_B]$, parameterize the viewing planes of spacecraft A and B. Then the intersection of the 2 planes yields the trajectory of the meteor as shown in Fig. 11. The trajectory line or the radiant direction is given by:

$$\xi_R = \frac{(b_A c_B - b_B c_A)}{d_R} \qquad \eta_R = \frac{(b_A c_B - b_B c_A)}{d_R} \qquad \zeta_R = \frac{(a_A b_B - a_B b_A)}{d_R} \qquad (3)$$

Where the length of the radiant vector is given by:

$$d_R = \sqrt{(b_A c_B - b_B c_A)^2 + (b_A c_B - b_B c_A)^2 + (b_A c_B - b_B c_A)^2} \qquad (4)$$

The radiant right ascensions and declinations are now found from spherical coordinates as:

$$\alpha_R = \tan^{-1}\frac{\eta_R}{\xi_R} \qquad\qquad \delta_R = \sin^{-1}\zeta_R \qquad (5)$$

With the radiant known the inertial location of $n^{th}$ pixel of the meteor event can be determined. According to [32], we can define the following relations:

The position vector of the event with respect to spacecraft A is given by:
$$X_{AN} = b_n c_A - c_n b_A \qquad Y_{AN} = c_n a_A - a_n c_A \qquad Z_{AN} = a_n b_A - b_n a_A \qquad (6)$$

The geocentric position of the $n^{th}$ pixel is then determined as:

$$X_N = X_{AN} + X_A \qquad Y_N = Y_{AN} + Y_A \qquad Z_N = Z_{AN} + Z_A \qquad (7)$$

Using this approach, we have shown how the position of the object seen by the two spacecraft can be calculated. The technique is then extended to calculating velocity and acceleration in a moving frame, together with the size of the object.

## 6. CONCLUSIONS

CubeSats thanks to their low developmental and launch costs offer new opportunities for rapidly demonstrating on-orbit surveillance capabilities. We propose the development of SWIMSat (Space based Wide-angle Imaging of Meteors) 3U CubeSat on-orbit demonstrator network that is designed to observe illuminated and unilluminated objects entering the Earth's atmosphere and transiting in its vicinity. We have presented a way to autonomously detect and track unilluminated objects with visual-light sensors. This method scales linearly with image size and is therefore very quick with respect to runtime complexity. Preliminary results show accuracy to less than ten pixels with a high detection rate. Importantly this object detection and tracking task can't be performed by humans. With this method, our SWIMSat orbital observatory network could contribute to the search for meteors and NEOs. While initially developed for CubeSats, this method may be of use to larger satellites or even ground-based observatories.

## 7. REFERENCES

1. J.M. Plane, "Cosmic dust in the earth's atmosphere," *Chemical Society Reviews* 41, pp. 6507-18, (2012).
2. P.G. Brown et al., "A 500-kiloton airburst over Chelyabinsk and an enhanced hazard from small impactors," *Nature* 503, 14 November 2013
3. S.D. Miller, W.C. Straka III, A. S. Bachmeier, T.J. Schmit, P.T. Partain, Y.-J. Noh, "Earth-viewing satellite perspectives on the Chelyabinsk meteor event," *PNAS*, 110 (45) 18092-18097, (2013).
4. S. Proud, "Reconstructing the orbit of the Chelyabinsk meteor using satellite observations," *Geophysical Resource Letters* (2013).
5. K. Hill, L. Rogers, R. Hawkes, "Sputtering and high altitude meteors," Earth, Moon, and Planets 95, pp. 403--12 (2004)
6. L. Rogers, K. Hill, R. Hawkes, "Mass loss due to sputtering and thermal processes in meteoroid ablation," Planetary and Space Science 53, pp. 1341-54 (2005).
7. D. Vinković, "Thermalization of sputtered particles as the source of diffuse radiation from high altitude meteors," Advances in Space Research 39, pp. 574-82 (2007).
8. T. Vondrak, J. Plane, S. Broadley, D. Janches, "A chemical model of meteoric ablation," Atmospheric Chemistry and Physics 8, pp. 7015-31, (2008).
9. S. Shin and W.-Y. Kim, "Fast satellite streak detection for high-resolution image," in 2018 International Workshop on Advanced Image Technology (IWAIT), pp. 1–4, IEEE, 2018.
10. J. McInroy, L. Robertson, and R. Erwin, "Autonomous distant visual silhouetting of satellites," IEEE Transactions on Aerospace and Electronic Systems, vol. 44, no. 2, 2008.
11. E. Murad, and I. P. Williams, eds. "Meteors in the Earth's Atmosphere: Meteoroids and Cosmic Dust and Their Interactions with the Earth's Upper Atmosphere," Cambridge University Press, (2002).
12. P. Jenniskens, S. J. Butow. "The 1998 Leonid multi-instrument aircraft campaign—an early review." Meteoritics & Planetary Science 34.6, pp. 933—943, (1999).
13. P. Jenniskens, S. J. Butow, M. Fonda. "The 1999 Leonid multi-instrument aircraft campaign—an early review." Leonid Storm Research. Springer, Dordrecht, pp.1-26, (2000).
14. J. Peter, W. R. Ray. "The 2001 Leonid Multi-Instrument Aircraft Campaign-an early review." The Institute of Space and Astronautical Science repot. SP 15 (2003).
15. P. Jenniskens, "The 2002 Leonid MAC airborne mission: first results." WGN, Journal of the International Meteor Organization 30 pp. 218-224, (2002).
16. V. Hernandez, P. Gankidi, A. Chandra, A. Miller, P. Scowen, H. Barnaby, E. Adamson, E., Asphaug, E., Thangavelautham, J., "SWIMSat: Space Weather and Meteor Impact Monitoring using a Low-Cost 6U CubeSat," Proceedings of the 30th Annual AIAA/USU Conference on Small Satellites, 2016.
17. V. Hernandez, A. Ravindran, M. Herreras-Martinez, E. Asphaug, J. Thangavelautham, "On-Orbit Demonstration of the Space Weather and Meteor Impact Monitoring Network," Proceedings of the 31st AIAA/USU Small Satellite Conference, 2017.
18. R. Nallapu, A. Ravindran, H. Kalita, V. Hernandez, V. Reddy, R. Furfaro, E. Asphaug, J. Thangavelautham, "Smart Camera System Onboard a CubeSat for Space-based Object Reentry and Tracking," IEEE/ION PLANS Conference, 2018.
19. E. Asphaug, J. Thangavelautham, "Asteroid Regolith Mechanics and Primary Accretion Experiments in a Cubesat," Proceedings of the 45th Lunar and Planetary Science Conference, 2014.
20. J. Lightholder, A. Thoesen, E. Adamson, J. Jakubowski, R. Nallapu, S. Smallwood, L. Raura, A. Klesh, E. Asphaug, J. Thangavelautham, "Asteroid Origins Satellite 1: An On-orbit CubeSat Centrifuge Science Laboratory," Acta Astronautica, Vol 133, pp. 81-94 (2017).

21. E. Asphaug, J. Thangavelautham, A. Klesh, A. Chandra, R. Nallapu, L. Raura, M. Herreras-Martinez, "A CubeSat Centrifuge for Long Duration Milligravity Research," Nature Microgravity, pp. 1-15, (2017).
22. J. T. Wright et al., "Exoplanet detection methods," in Planets, Stars and Stellar Systems, pp. 489–540, Springer, 2013.
23. M. Zuliani, "Ransac for dummies," Vision Research Lab, University of California, Santa Barbara, 2009.
24. J. Vourvoulakis, J. Lygouras, and J. Kalomiros, "Acceleration of ransac algorithm for images with affine transformation," in Imaging Systems and Techniques (IST), 2016 IEEE International Conference on, pp. 60–65, IEEE, 2016.
25. J. P. Balarini and S. Nesmachnow, "A c++ implementation of otsus image segmentation method," Image Processing On Line, vol. 6, pp. 155–164, 2016.
26. A. Rushby, "A multiplicity of worlds: Other habitable planets," Significance, vol. 10, no. 5, pp. 11–15, 2013.